\def\year{2022}\relax
\documentclass[letterpaper]{article} 
\usepackage{aaai22}  
\usepackage{times}  
\usepackage{helvet}  
\usepackage{courier}  
\usepackage{appendix}
\usepackage[hyphens]{url}  
\usepackage{graphicx} 
\usepackage{natbib}  
\usepackage{caption} 
\DeclareCaptionStyle{ruled}{labelfont=normalfont,labelsep=colon,strut=off} 
\usepackage{algorithm}
\usepackage{algorithmic}

\usepackage{newfloat}
\usepackage{listings}
\floatstyle{ruled}
\newfloat{listing}{tb}{lst}{}
\floatname{listing}{Listing}
\usepackage{subfigure}
\usepackage{booktabs}
\usepackage{multirow}
\usepackage{soul}
\usepackage{amsmath}
\usepackage{amssymb}

\title{Hybrid Graph Neural Networks for Few-Shot Learning}

\author {
        Tianyuan Yu\textsuperscript{\rm 1,2},
        Sen He\textsuperscript{\rm 1,3},
        Yi-Zhe Song\textsuperscript{\rm 1,3},
        Tao Xiang\textsuperscript{\rm 1,3}\\
}
\affiliations {
    \textsuperscript{\rm1}Center for Vision, Speech and Signal Processing, University of Surrey  \\
    \textsuperscript{\rm2}National University of Defense Technology  \\
    \textsuperscript{\rm3}iFlyTek-Surrey Joint Research Centre on Artificial Intelligence  \\
    \{tianyuan.yu, sen.he, y.song, t.xiang\}@surrey.ac.uk
}

\begin{document}

\maketitle
\begin{abstract}
Graph neural networks (GNNs) have been used to tackle the  few-shot learning (FSL) problem and shown great potentials under the transductive setting. However under the inductive setting, existing GNN based methods are less competitive. This is because they use an instance GNN as a label propagation/classification module, which is jointly meta-learned with a feature embedding network. This design is problematic because the classifier needs to adapt quickly to new tasks while the embedding does not. To overcome this problem, in this paper we propose a novel hybrid GNN (HGNN) model consisting of two GNNs,  an instance GNN and a prototype GNN. Instead of label propagation, they act as  feature embedding adaptation modules for quick adaptation of the meta-learned feature embedding to new tasks. Importantly they are designed to deal with a fundamental yet often neglected challenge in FSL, that is, with only a handful of shots per class, any few-shot classifier would be sensitive to badly sampled shots which are either outliers or can cause inter-class distribution overlapping. 
 Extensive experiments show that our HGNN obtains new state-of-the-art on three FSL benchmarks. The code and models are available at  \textit{https://github.com/TianyuanYu/HGNN}.
\end{abstract}

\section{Introduction}

Deep convolutional neural networks (CNNs) have achieved  great successes in various computer vision problems including  image classification~\cite{krizhevsky2012imagenet}, semantic segmentation \cite{chen2017deeplab}, object detection \cite{ren2015faster} and image captioning \cite{xu2015show}. However, training deep neural networks requires a large amount of labeled data (e.g., hundreds of samples per class). Collecting and annotating them is often tedious, expensive and even infeasible for some rare classes.  This thus hinders their deployments in real-world applications.  One widely studied solution to this problem is few-shot learning (FSL)~\cite{vinyals2016matching,finn2017model,snell2017prototypical,sung2018learning,sun2019meta,zhang2020deepemd}, which aims to recognize a set of novel classes with only a handful  of labeled samples/shots (e.g., 1-5 per class) by knowledge transfer from a set of base classes with abundant samples.


\begin{figure}[t]
\begin{center}
  \includegraphics[width=0.45\textwidth]{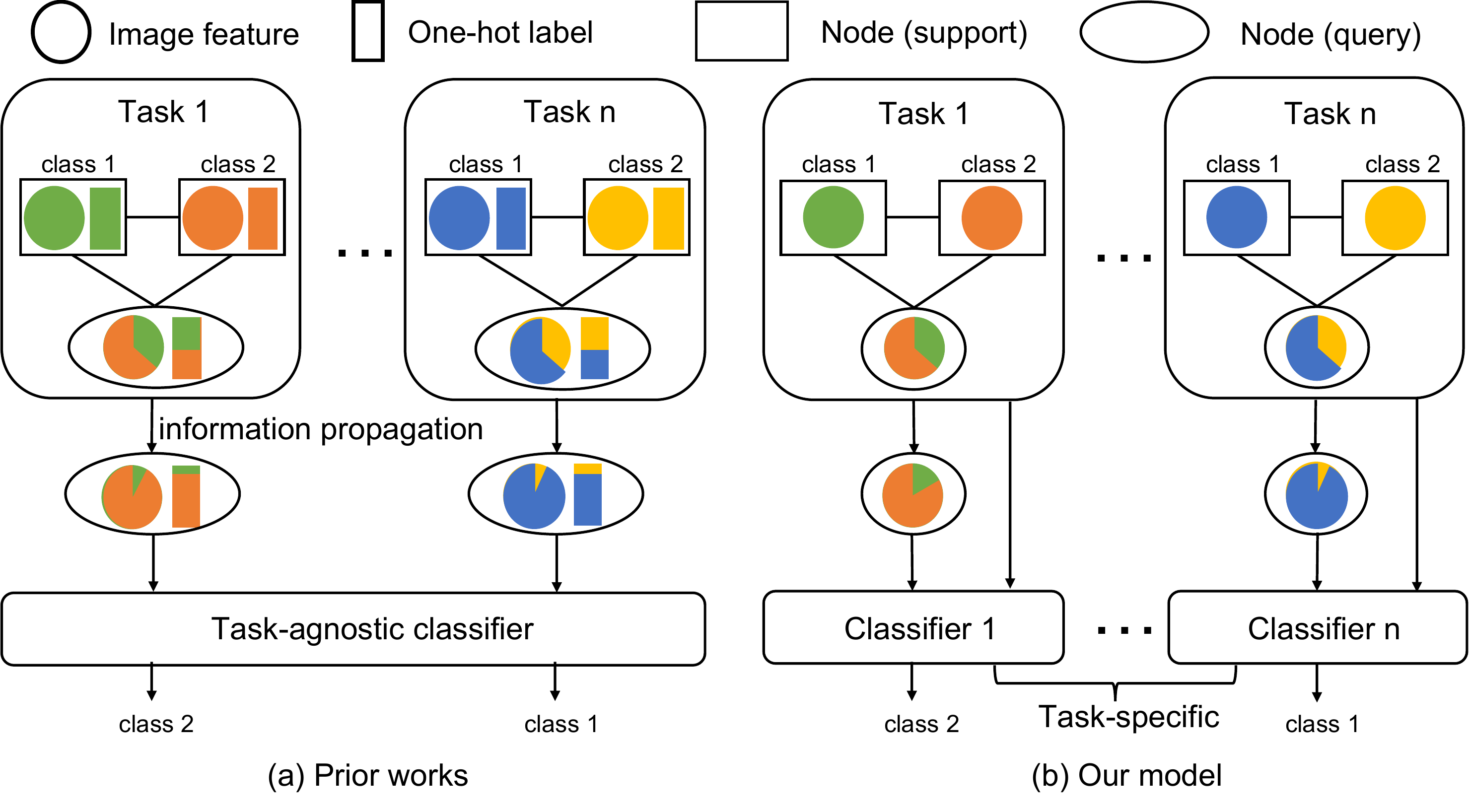}
\end{center}
\vspace{-0.4cm}
\caption{\small  Illustration of the differences between our GNN and prior GNN in FSL using  2-way 1-shot tasks. (a) Previous methods, e.g. \cite{garcia2017few},  use GNN for label propagation, i.e., as a task-agnostic classifier. (b) We use GNN for feature adaptation and leave the query image label prediction job to  task-specific classifiers. }
\label{fig:difference}
\vspace{-0.7cm}
\end{figure}

Most recent FSL approaches follow the paradigm of meta-learning \cite{hospedales2020metalearning} with episodic training. Concretely, a model is trained  in each episode with a few-shot classification task sampled from the base classes. Each task consists of a support set and a query set for inner and outer loop training respectively. This is to imitate the meta-test setting, under which only few labeled data are given for a novel task. The objective is to meta-learn  a model capable of ``learning to learn", that is, generalizing well to new tasks composed of unseen classes. Existing approaches differ primarily on what is meta-learned -- a deep embedding/distance metric \cite{vinyals2016matching,snell2017prototypical,sung2018learning,zhang2020deepemd} or an optimization algorithm \cite{finn2017model}.

Among various existing FSL approaches,  graph neural network (GNN) \cite{kipf2016semi} based FSL methods \cite{garcia2017few,luo2019learning,liu2019TPN,kim2019edge,yang2020dpgn} have received increasing attention due to their excellent performance under the transductive setting. These methods, as shown in Figure 1(a), employ GNN as a label propagation module whereby the graph is used for either node label prediction \cite{garcia2017few,liu2019TPN,luo2019learning} or edge label prediction \cite{kim2019edge,yang2020dpgn}. In other words, a GNN is used  as a classifier which takes a feature embedding network's output as input and produces class labels. Both the classifier/GNN parameters and the feature embedding network parameters are learned jointly in the outer loop as two parts of a single model. According to~\cite{garcia2017few,kim2019edge}, a GNN is naturally suited for few-shot learning due to its ability to aggregate knowledge by message passing on a graph constructed on the limited support set instances as well as the query set instances. However, the efficacy of the existing methods \cite{garcia2017few,luo2019learning,liu2019TPN,kim2019edge,yang2020dpgn} 
under the inductive setting is still lagging behind the state-of-the-art \cite{zhang2020deepemd,ye2020few}. 

\begin{figure}[t]
\begin{center}
\subfigure[Feature embedding]{
\includegraphics[width=.22\textwidth]{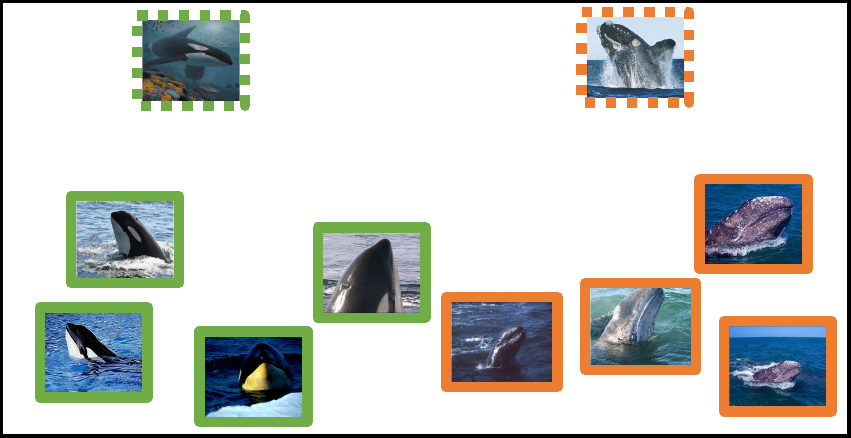}  \label{fig:concept_a}
}
\subfigure[HGNN-adapted embedding]{
\includegraphics[width=.22\textwidth]{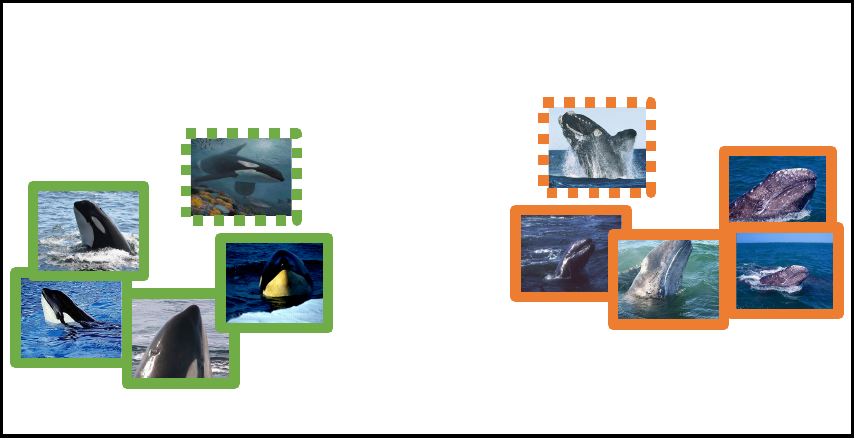}\label{fig:concept_b}
}
\end{center}
\vspace{-0.4cm}
\caption{ (a) Illustration of two issues, i.e., intra-class outliers and inter-class overlapping,  caused by badly sampled instances in  2-way 5 shot tasks. (b) With our HGNN, the outlier samples' effects are minimized and the two classes become well separated. More illustration with real data distributions can be found in Figure \ref{fig:visualisation}.}
\vspace{-0.4cm}
\label{fig:concept}
\end{figure}

We believe that it is the joint meta-learning of the classifier and feature embedding that impedes the effectiveness of existing GNN-based FSL methods under the inductive setting. Specifically, there is an on-going debate \cite{raghu2019rapid,oh2020does,tian2020rethinking} on what meta-learning is truly about: rapid learning or feature reuse, or both? There seems to be little doubt on the importance of learning a good feature embedding, to the extent that it was argued recently that a good embedding is all one needs \cite{tian2020rethinking}. Meanwhile, the ability to rapid adaptation to the new task at hand can  also be critical \cite{ye2020few}. Nevertheless, there is an emerging consensus that jointly meta-learning  both the classifier which has to be adapted quickly to each new task, and the feature embedding network which is intrinsically task-agnostic for feature reuse, is perhaps not a good idea due to their contradictory nature \cite{raghu2019rapid,oh2020does}.

To overcome this limitation, in this paper, a novel \textbf{\textit{Hybrid Graph Neural Network}} (HGNN) based FSL framework is proposed. As shown in Figure 1(b), different from existing GNN-FSL methods \cite{garcia2017few,luo2019learning,liu2019TPN,kim2019edge,yang2020dpgn} whereby GNNs are used as classifiers via label propagation from support to query, our GNNs are used as  feature embedding task adaptation modules to address a specific problem that is often ignored in FSL. That is,  when only few support set instances are available  to represent  each class in a support set, any classifier built on them would be sensitive to badly drawn samples. More specifically, as shown in Figure \ref{fig:concept_a}, two issues can be caused by bad samples: outliers and inter-class overlapping.  Outliers, caused by unusual pose/lighting etc \cite{yu2019robust}, are problematic for any learning tasks, but particular so in FSL -- with few samples per class, a single outlier could have an immense effect on the class distribution. The issue of inter-class overlapping is also commonplace when the training samples of different classes have very similar background or object pose, or just being visually similar. Through message passing across the whole supports set containing all classes, our GNNs are learned to identify these outliers, minimize their negative effects, and re-adjust the class distributions so that each class' distribution is compact and further apart from each other (see Figure \ref{fig:concept_b}).  

Concretely, our HGNN is integrated with a feature embedding network meta-learned using the popular prototypical network (ProtoNet) \cite{snell2017prototypical}. As shown in Figure \ref{fig:overview}, it consists of two GNNs, namely an \textbf{\textit{Instance Graph Neural Networks}} (IGNN) constructed with the whole support set samples/instances plus a single query sample as nodes, and a \textbf{\textit{Prototypical Graph Neural Network}} (PGNN) whose nodes correspond to class prototypes. They are designed to address the two bad sampling issues respectively. In particular, the IGNN focuses on outlier identification and neutralization through instance-level message passing, while the PGNN operates at the class prototype level to make sure that different classes are well separable in the embedding space adapted by the GNN. These two objectives are clearly complementary and our HGNN exploits this using two graph-specific losses and an inter-graph consistency loss.    

Our contributions are as follows: (1) We propose a new framework for using GNNs for FSL which differs from existing GNN-FSL methods in that it uses the GNNs for feature embedding adaptation for new tasks, rather than label prediction. (2)  As an instantiation, we propose an instance GNN which is designed to address the outlying sample problem. (3) We further propose a prototype GNN to deal with overlapping classes, which, as far as we know, has never been used before for FSL. (4) These two GNNs are integrated into a hybrid graph model which produces new state-of-the-art on three widely used FSL datasets.

\section{Related Work}

\subsection{Meta-learning}

Most FSL methods are based on meta-learning, which aims to learn, through episodic training a model that can generalize to unseen new tasks.  Depending on what is meta-learned, existing methods can be divided broadly into two categories, {\textit{optimization-based }} 
and \textit{metric-based}. 

\noindent \textbf{{Optimization-based Methods}} target at learning a good optimization algorithm that can adapt a deep CNN to a new task represented with few samples. Early works focused on meta-learning an optimizer. These include the LSTM-based meta-learner \cite{Ravi2017OptimizationAA} and the external-memory assisted weight updating \cite{MunkhdalaiY17}.  Later, the focus shifted to meta-learning a model initialization suitable for fast adapting the model to a new task by fine-tuning  on  few support samples with few iterations. The representative works are MAML~\cite{finn2017model} and its many variants \cite{DBLP:journals/corr/abs-1803-02999,rajeswaran2019iMAML}. These methods are faced with a difficult bi-level optimization problem due to the inter-dependency of the parameters updated in the inner and outer loops in each episode. To overcome this challenge,  \cite{lee2019meta} used implicit differentiation for quadratic programs in a final SVM layer and \cite{sun2019meta} proposed to learn task-relevant scaling and shifting functions to dynamically adjust the CNN weights. 

\noindent \textbf{{Metric-based Methods}} aim to learn a transferable feature embedding or distance metric. In the early works, MatchingNet~\cite{vinyals2016matching} used an attention mechanism over the learned representations and applied nearest neighbor search for classification. ProtoNet~\cite{snell2017prototypical} used the mean of each class's support set in the embedding space as a prototype without any classifier parameter and meta-learned the feature embedding network in the outer loop using a query set. RelationNet~\cite{sung2018learning} learned a distance metric  through a neural network on top of a feature embedding network. Many recent works, similar to our HGNN, were formulated with the ProtoNet~\cite{snell2017prototypical} as basis, due to its simplicity and competitive performance. For example, \cite{allen2019infinite} represented each class as a set of clusters rather than a single cluster. \cite{li2019revisiting} replaced the global feature by a local descriptor. \cite{afrasiyabi2019associative} introduced the idea of associative alignment for leveraging each informative part of the support data. Most existing works used holistic image features. In contrast, DeepEMD \cite{zhang2020deepemd} showed that image-patch features can be useful in addressing the spatial mis-alignment issue. 


\noindent \textbf{{Rapid Learning vs Feature Reuse}} Despite their theoretical attractiveness, optimization-based methods are in general less competitive compared to metric-based methods. This triggered discussions on the merits of rapid learning and feature reuse \cite{raghu2019rapid,oh2020does,tian2020rethinking}. The optimization-based methods obviously focus on rapid learning, while metric-based methods are mostly about feature reuse. Yet, it was discovered that even those optimization-based methods also  heavily rely on learning a good stable feature embedding that can be used for any new task \cite{raghu2019rapid,oh2020does}. Based on this understanding, it was suggested that the classifier parameters, which must be adapted rapidly to new tasks, should not be meta-learned jointly with the feature embedding parameters, a principle adopted by our HGNN.    Recently, \cite{tian2020rethinking} suggested that feature embedding learning is all one needs - pre-training a feature embedding network together with model distillation can bypass the episodic training stage altogether.  A similar finding was reported in \cite{liu2020negative}. However, it was argued that a task adaptation module (not a classifier), jointly learned with a feature embedding, is the best trade-off between rapid learning and feature reuse \cite{oreshkin2018tadam,ye2020few}. Our HGNN provides novel task adaptation modules in the form of both instance and prototype GNNs and yields clearly better results than \cite{oreshkin2018tadam,ye2020few}.

\subsection{Graph Neural Networks in Few-Shot Learning}

Garcia \textit{et al.} ~\cite{garcia2017few} were the first to use GNNs to address few-shot classification tasks. In their GNNs, each node corresponds to one instance (support or query) and is represented as the concatenation of a feature embedding and a label embedding. 
The final layer in their model is a linear classification layer which directly outputs the prediction scores for each query node. 
\cite{liu2019TPN} proposed to learn to propagate labels from support nodes to query nodes under the transductive setting, by learning a graph construction module that exploits the manifold of data in a latent space. 
Similarly focusing on the transductive setting but different from these node label prediction modules, EGNN \cite{kim2019edge} learned to predict the edge-labels on the graph. Based on EGNN, \cite{luo2019learning} jointly modeled the long-term inter-task correlations and short-term intra-class adjacency with the derived continual graph neural networks, which can retain and then access important prior information associated with newly encountered episodes.   Recently,  Yang \textit{et al.}~\cite{yang2020dpgn} proposed DPGN, a dual graph network consisting of a point graph and a distribution graph, in which each instance node  is used to combine the distribution-level and instance-level relations.  

Most of these GNN based FSL methods focus on the transductive setting, under which the full test query set can be injected into the graph for label propagation to alleviate the lack of training sample problem. However, their inductive setting performance is lagging behind the state-of-the-art. As mentioned early, we hypothesize that this is because label propagation means that these GNNs are essentially classifiers, and jointly meta-learning a classifier and a feature embedding confuses the model on whether to emphasize on rapid learning or feature reuse. In contrast, our HGNN removes the label propagation functionality and focuses on feature embedding task adaptation. Further, different from these instance GNN only methods, we additionally introduce a prototype GNN. As a result, our HGNN produces the new state-of-the-art under the inductive setting on several benchmarks.  

\vspace{-0.2cm}

\section{Methodology}

\begin{figure*}[t]
\begin{center}
  \includegraphics[width=0.9\textwidth]{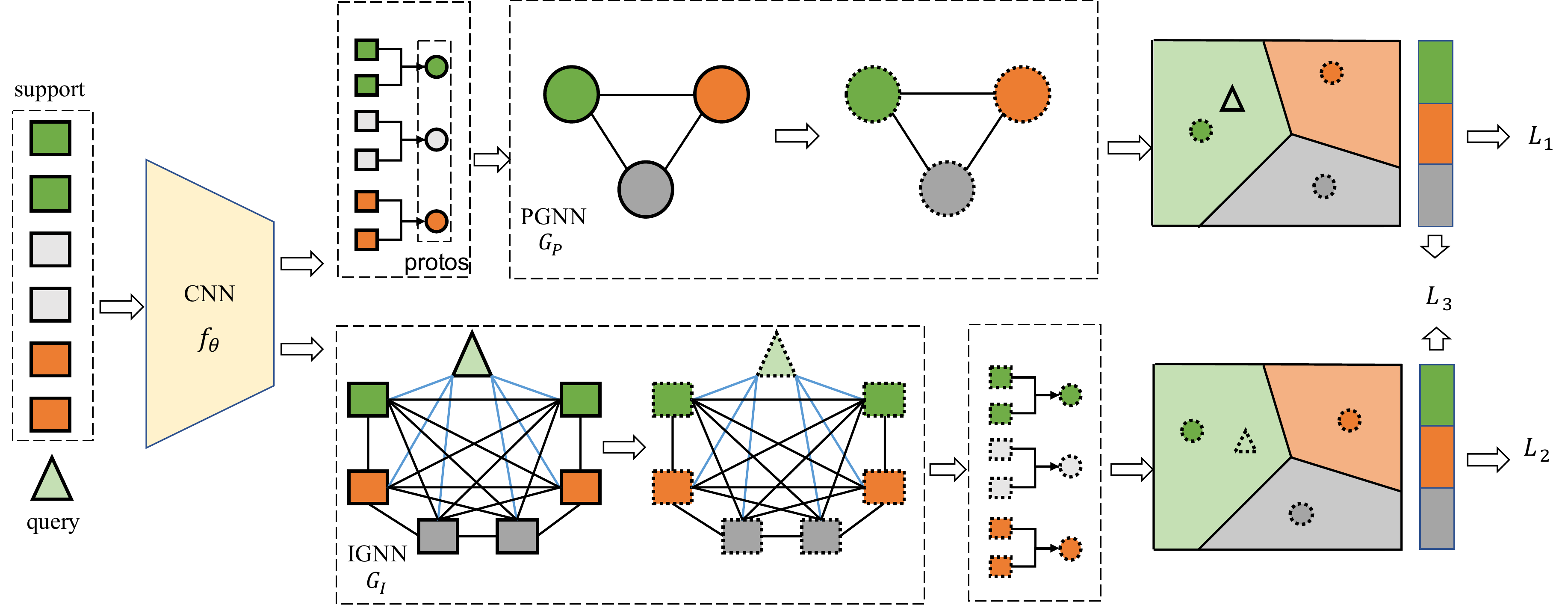}
\end{center}
\caption{\small Overview of our HGNN model in a 3-way 2-shot case. Features extracted by a feature embedding CNN are fed into a PGNN and an IGNN for task adaptation. In the PGNN, each node represents a class prototype, which is initialized by averaging the support set features for that class. The nodes of the IGNN, on the other hand, include all the instances in the support set as well as a single query instance. After instance feature adaptation using the IGNN, the updated instance features are used to produce another set of class prototypes. Note that in the IGNN, the edge/message passing from instances in the support set to the query instance is one-directional from support to query  (blue),  while the edges between instances in the support set are bidirectional (black). These two sets of GNN-updated prototypes are  used to predict the class of the query instances and the predictions are evaluated using cross-entropy losses ($L_1$ and $L_2$), and a consistency loss ($L_3$) is used to enforce the prediction consistency between the two GNNs. 
}
\vspace{-0.6cm}
\label{fig:overview}
\end{figure*}

\subsection{Problem Definition}
We follow the standard FSL formulation \cite{vinyals2016matching,finn2017model,snell2017prototypical}. Concretely,   a task is a  N-way K-shot classification problem sampled from a meta-test set $\mathcal{D}_{\text{TST}}$. Adopting an episodic-training based meta-learning strategy, N-way K-shot tasks are sampled from a meta-training dataset $\mathcal{D}_{\text{TRN}}$, in order to imitate the few-shot test setting. Note that there is no overlapping between the class label spaces of the two sets, i.e., $\mathcal{C}_{\text{TRN}} \cap \mathcal{C}_{\text{TST}} = \varnothing$. In each episode, we first sample N classes $\mathcal{C}_{N}$ from the training class space randomly.  Training instances are then sampled from these classes to form a support set $\mathcal{S}_{N}$ and a query set $\mathcal{Q}_{N}$ consisting of K and Q samples from each sampled class in $\mathcal{C}_{N}$ respectively. The sampled training instances are denoted as $\mathcal{S}_{N}=\{(x_{i},y_{i})~|~y_{i}\in \mathcal{C}_{N},i = 1,\dots, N \times K\}$ and $\mathcal{Q}_{N}=\{(x_{i},y_{i})~|~y_{i}\in \mathcal{C}_{N},i = 1,\dots, N \times Q\}$, where $\mathcal{S}_{N} \cap \mathcal{Q}_{N} = \varnothing$. During training, the model uses the support set $\mathcal{S}_{N}$ to build a classifier in an inner loop, and then the query set $\mathcal{Q}_{N}$ is used in an outer loop to evaluate and update the model parameters.

\subsection{Label Propagation GNNs  for FSL}

Before introducing our framework, we first briefly describe the formulation of existing GNN-based FSL methods in order to highlight the differences in our design. As mentioned earlier, all existing GNN-based FSL methods \cite{garcia2017few,luo2019learning,liu2019TPN,kim2019edge,yang2020dpgn} use a GNN as a label propagation module, i.e., a label classifier. To explain this, we use the model in   
\cite{garcia2017few} as an example. {In this model, the GNN is a fully-connected graph composed of $M$ nodes. Each node represents an instance from either a support set or a query set. Under the inductive setting, only one query instance is included in the graph.  Therefore for a $N$-way $K$-shot task, we have $M=N\times K+1$. In contrast, all query instances are used to construct the graph under the transductive setting; we thus have $M=N\times(K+Q)$. } For a graph $G$ with L layers, the input with M entries from the $l^{\text{th}}$ layer is denoted as $F^{l} \in \mathbb{R}^{
M\times (d_{f}+d_{l})}$, where $d_{f}$ is the dimension of the instance feature obtained from a feature embedding network, and $d_{l}$ is the dimension of label embedding. In other words, each node is represented as a concatenation of a visual feature embedding and a label embedding indicating which class each support set instance belongs to.  The output of the $(l+1)^{\text{th}}$ layer is given by
\begin{equation} \label{eq:1}
    F^{l+1} = G^{l}(F^{l})= \rho \left(\mathcal{A}^{l}\phi^{l} (F^{l})\right),
\end{equation}
where $\rho$ is an element-wise non-linear activation function, $\phi^{l}$ is a linear transformation layer 
, and $\mathcal{A}^{l} \in \mathbb{R}^{M \times M}$ is an adjacency operator in layer $l$. Each entry in $\mathcal{A}^{l}$ is computed as
\begin{equation}
    \mathcal{A}^{l}_{ij} = \psi(\phi^{l}(F_{i}),\phi^{l}(F_{j})),
    \label{eq:adjacency}
\end{equation}
where $\psi$ is a neural network. $\mathcal{A}^{l}$ is normalized along each row.

The final class prediction score of a query node is computed as
\begin{equation}
    \mathbf{p}_{i} = F^{L}_{i}\mathbf{w},
\end{equation}
where $\mathbf{w} \in \mathbb{R}^{(d_{f}+d_{l})\times N}$ parameterizes a linear classification layer.

It is clear from the formulation of this GNN that, its objective is to take the support set labels as part of the node representation at the input layer of the graph and perform label propagation on the graph. As a result, the query sample nodes at the output layer can be used directly for label prediction. The alternative designs in \cite{kim2019edge, luo2019learning,yang2020dpgn} encode the support set labels on the edges, instead of nodes of the graph, but the main functionality of the GNN as a label classifier remains the same. In contrast, our HGNN does not encode the label information anywhere in the graph and it serves as a feature embedding task adaptation module as described in detail next.   

\subsection{Hybrid GNN for FSL}
The proposed GNN based FSL framework is illustrated in Figure~\ref{fig:overview}. It consists of a feature extraction backbone and a hybrid graph neural network (HGNN) that is further composed of an instance GNN (IGNN) and a prototype GNN (PGNN).  In each training episode, we extract the features for the images in the support set $\mathcal{S}_{N}$ and the query set $\mathcal{Q}_{N}$ using a feature embedding CNN. The extracted features are then fed into the GNNs as node features for task adaptation. A key difference between  the proposed GNNs and the prior ones in \cite{garcia2017few,luo2019learning,liu2019TPN,kim2019edge,yang2020dpgn} is that our GNNs contain no label information. 
\subsubsection{Prototypical Graph Neural Network (PGNN)}
As illustrated in Figure \ref{fig:concept_a}, as each class in the support set $\mathcal{S}_{N}$ is only represented by $K$ samples, a FSL model is challenged by two issues caused by badly sampled instances, namely outliers and class overlapping.  Our PGNN is designed to address the class overlapping problem. More specifically, since our HGNN is integrated with a ProtoNet  FSL model where each class in $\mathcal{S}_{N}$ is represented by the class mean or prototypes, we feed the prototypes into the PGNN and use message passing between them to ensure that class overlapping is minimized.

Formally, as shown in Figure \ref{fig:overview},  our PGNN $G_{P}$ receives the prototypes' features $F_{P} \in \mathbb{R}^{N\times d}$  of N classes in the support set as nodes' features, where $F_{P_{n}}  = \frac{1}{K} \sum_{i=1}^{K}f_{\theta}(x_{n_i})$, where $f_{\theta}(\cdot)$ is a feature embedding network producing features of $d$ dimensions, and $x_{n_i}$ is the $i^{\text{th}}$ image from the $n^{\text{th}}$ class in the support set. To stabilize the training, as per standard, we adopt the residual connection \cite{he2016deep} and the layer norm \cite{ba2016layer} in our GNNs. Thus the output of an one-layer PGNN is computed as
\begin{equation}
    \hat F_{P} = \text{LayerNorm}(F_{P}+\varphi_P(G_{P} (F_{P}))),
\end{equation}
where $\varphi$ is a linear transformation layer {to improve the expressive power of the adapted  features with the same dimension $d$,} and $G_{P}$ denotes the same operations in Equation~\ref{eq:1}. Finally, the refined prototypes are used to classify the query samples in $\mathcal{Q_{N}}$. Concretely,  {the probability of the $i$th query belonging to the class $j$ is first computed as}
\begin{equation}
    p^{P}_{i,j} = \frac{\text{exp}(-Ed(f_{\theta}(x_{i}),\hat F_{P_{j}}))}{\sum_{k=1}^{N}\text{exp}(-Ed(f_{\theta}(x_{i}),\hat F_{P_{k}})},
\end{equation}
where $Ed(\cdot, \cdot)$ is the Euclidean distance. To maximize the probability $p^{P}_{i,j}$, the learned PGNN has the incentive to re-arrange the relative position of the $N$ prototypes so that they become more separable. This way, it becomes easier to assign each query image to the correct class with high confidence.

\subsubsection{Instance Graph Neural Network (IGNN)}
The PGNN focuses on the inter-class relationship with the class mean or prototypes as graph node. It thus has limited ability to deal with the outliers that are best identified when intra-class instance relationships are examined. To that end, an IGNN is formulated.   

Since we focus on  the inductive setting, that is, only one query instance is available for inference at a time,  our IGNN consists of the whole support set instances and one query set instance as nodes (see Figure \ref{fig:overview}). For a N-way K-shot task, there are $N\times K+1$ nodes in the graph. This means that for $Q$ query set samples in a training episode, $Q$ graph needs to be constructed. Formally, the $i^{\text{th}}$ instance graph takes the $i^{\text{th}}$ query raw feature together with all support set samples' feature extracted by the feature embedding network $f_{\theta}(\cdot)$ as the nodes $F_{I_{i}} \in \mathbb{R}^{(N\times K+1)\times d}$. Similar to the PGNN, the features of nodes in our IGNN are refined by:

\begin{equation}
    \hat F_{I_{i}} = \text{LayerNorm}(F_{I_{i}}+\varphi_I(G_{I} (F_{I_{i}}))).
\end{equation}

Note that the updated nodes include all support nodes and a single query node. However, as shown in Figure \ref{fig:overview}, the message passing between support and query is one-directional (from support to query only) and the query has no effect on the support set node updating. This means that though we have Q graphs, the support set instances only need to be updated once and the additional computation is only on the query set instances. As a result, once trained the inference on our IGNN is very efficient. 

With the updated support set features, to adhere to the ProtoNet pipeline,  we compute another set of prototypes for each class by computing the updated support set instance class means. Finally, the probability of the $i^{\text{th}}$ query node  belonging to class $j$ is
\begin{equation}
    p^{I}_{i,j} = \frac{\text{exp}(-Ed(\hat F_{I_{iq}}, \Tilde{F_{I_{iPj}}}))}{\sum_{k=1}^{N}\text{exp}(-Ed(\hat F_{I_{iq}},\Tilde{F_{I_{iPk}}}))},
\end{equation}
where $\hat F_{I_{iq}}$ is the updated query node feature in the $i^{\text{th}}$ IGNN, and $\Tilde{F_{I_{iPj}}}$ is the prototype for the $j^{\text{th}}$ class. To maximize the probability $p^{I}_{i,j}$, the learned IGNN is encouraged to identify the outlying support set samples and use the other instances of the same class to `pull' it closer in the updated embedding space. Together with PGNN, IGNN can create a  more friendly embedding space for classifying a query sample by comparing it with the support set samples. This is illustrated in Figure \ref{fig:concept_b} and verified both quantitatively and qualitatively in our experiments.   

\subsection{Training Objectives}

During training, the two GNNs in our HGNN make predictions for each query using their respective prototypes, and trained together with the shared feature embedding network $f_{\theta}$ via cross entropy losses. All the parameters in  $f_{\theta}$, PGNN $G_P$, and IGNN $G_I$ are end-to-end trained.

Specifically, the classification losses on PGNN and IGNN are
\begin{equation}
L_1 = \sum_i^Q\sum_j^{N}-\mathbb{I}(y_i==j)\operatorname{log}(p_{i,j}^P),
\label{eq:l1}
\end{equation}
\begin{equation}
L_2 = \sum_i^Q\sum_j^{N}-\mathbb{I}(y_i==j)\operatorname{log}(p_{i,j}^I),
\label{eq:l2}
\end{equation}
where $y_i$ is the ground truth label of the $i^{th}$ query and $\mathbb{I}(x)$ is an indicator function: $\mathbb{I}(x)=1$ when $x$ is true and 0 otherwise.

In addition, to make the prediction scores for each query from the GNNs to be consistent in our HGNN, a symmetric Kullback–Leibler (KL) divergence loss is used:
\begin{equation}
\begin{aligned}
L_3= \sum_j^N p_{i,j}^I\operatorname{log}\frac{p_{i,j}^I}{p_{i,j}^P}+ \sum_j^N p_{i,j}^P \operatorname{log} \frac{p_{i,j}^P}{p_{i,j}^I}.
\end{aligned}
\label{eq:l3}
\end{equation}

Thus, during training, the total loss is:

\begin{equation}
L(\theta, \phi_I, \phi_P) = L_1 + L_2 + L_3.
\label{eq:l}
\end{equation}

During meta-test, the class prediction of a query is given by the mean of two prediction scores from the two GNNs in our HGNN.

\section{Experiments}

\begin{table}[t]
\centering
\small\addtolength{\tabcolsep}{-2pt}
\caption{5-way 1/5-shot classification accuracy (\%) and 95\% confidence interval on \textit{Mini}ImageNet. $*$ indicates our reproduced results with the same pre-trained backbone, and $\dagger$ means transductive setting. }
\label{tab:mini}
\begin{tabular}{lccc}
\toprule
Method      & Backbone & 1-shot     & 5-shot     \\ \midrule
ProtoNet$^{*}$   & Conv4    & 52.78 $\pm$ 0.45 & 71.26 $\pm$ 0.36 \\
MAML        & Conv4    & 48.70 $\pm$ 1.84 & 63.10 $\pm$ 0.92 \\
Centroid    & Conv4    & 53.14 $\pm$ 1.06 & 71.45 $\pm$ 0.72 \\
Neg-Cosine  & Conv4    & 52.84 $\pm$ 0.76 & 70.41 $\pm$ 0.66 \\
FEAT        & Conv4    & 55.15 $\pm$ 0.20 & 71.61 $\pm$ 0.16 \\
\midrule
GNN$^{*}$        & Conv4    & 52.21 $\pm$ 0.20 & 67.03 $\pm$ 0.17 \\
EGNN        & Conv4    & 51.65 $\pm$ 0.55 & 66.85 $\pm$ 0.49 \\
EGNN$^{*}$        & Conv4    & 48.99 $\pm$ 0.59 & 61.99 $\pm$ 0.43 \\
TPN $^{\dagger}$ & Conv4    & 53.75 $\pm$ n/a  & 69.43 $\pm$ n/a \\
BGNN$^{*}$      & Conv4    &  52.35 $\pm$ 0.42 &  67.35 $\pm$ 0.35 \\
DPGN$^{*}$      & Conv4    & 53.22 $\pm$ 0.31 & 65.34 $\pm$ 0.29  \\ \midrule
HGNN        & Conv4    &     \textbf{55.63 $\pm$ 0.20}       & \textbf{72.48 $\pm$ 0.16} \\ \midrule \midrule
ProtoNet$^{*}$    & ResNet-12    & 62.41 $\pm$ 0.44 & 80.49 $\pm$ 0.29 \\
Neg-Cosine      & ResNet-12    & 63.85 $\pm$ 0.81 & 81.57 $\pm$ 0.56 \\
Distill      & ResNet-12    & 64.82 $\pm$ 0.60 & 82.14 $\pm$ 0.43 \\
DSN-MR    & ResNet-12    & 64.60 $\pm$ 0.72 & 79.51 $\pm$ 0.50 \\
DeepEMD  & ResNet-12    & 65.91 $\pm$ 0.82 & 82.41 $\pm$ 0.56 \\
FEAT       & ResNet-12    & 66.78 $\pm$ 0.20 & 82.05 $\pm$ 0.14 \\

E$^{3}\text{BM}$       & ResNet-25    & 64.3 $\pm$ n/a  & 81.0 $\pm$ n/a \\
MABAS        & ResNet-12    & 65.08 $\pm$0.86  & 82.70 $\pm$0.54 \\ 
APN       & CapsuleNet   & 66.43 $\pm$0.26  & 82.13 $\pm$0.23 \\
MetaOptNet+ArL       & ResNet-12   & 65.21 $\pm$0.58  & 80.41 $\pm$0.49 \\
PSST       & WRN-28-10   & 64.16 $\pm$0.44  & 80.64 $\pm$0.32 \\
FRN       & ResNet-12   & 66.45 $\pm$0.19  & 82.83 $\pm$0.13 \\
 \midrule
HGNN       & ResNet-12    &   \textbf{67.02 $\pm$ 0.20}        & \textbf{83.00 $\pm$ 0.13} \\ \bottomrule
\end{tabular}
\vspace{-0.4cm}
\end{table}

\begin{table}[t]
\small\addtolength{\tabcolsep}{-2pt}
\caption{Results on \textit{Tiered}ImageNet}
\label{tab:tiered}
\begin{tabular}{lccc}
\toprule
Method                                                            & Backbone & 1-shot                                    & 5-shot                                     \\\midrule
ProtoNet$^{*}$                                             & Conv4    & 50.89 $\pm$ 0.21      & 69.26 $\pm$ 0.18       \\
MAML                                                            & Conv4    & 51.67 $\pm$ 1.81      & 70.30 $\pm$ 0.08       \\
E$^{3}\text{BM}$    & Conv4 & 52.1 $\pm$ n/a                                      & 70.2 $\pm$ n/a  \\ \midrule
GNN$^{*}$                     & Conv4 &                 42.37 $\pm$ 0.20        & 62.54 $\pm$ 0.19                           \\
EGNN$^{*}$        & Conv4    & 47.40 $\pm$ 0.43 & 62.66 $\pm$ 0.57 \\
BGNN$^{*}$        & Conv4    & 49.41 $\pm$ 0.43  & 65.27 $\pm$ 0.35 \\
DPGN$^{*}$       & Conv4    & 53.99 $\pm$ 0.31 &  69.86$\pm$    0.28                     \\ \midrule
HGNN                                                              & Conv4    & \textbf{56.05 $\pm$ 0.21}      & \textbf{72.82 $\pm$ 0.18}       \\ \midrule
ProtoNet$^{*}$ & ResNet-12 & 69.63 $\pm$ 0.53                          & 84.82 $\pm$ 0.36                           \\
Distill        & ResNet-12 & 71.52 $\pm$ 0.69                          & 86.03 $\pm$ 0.49                           \\
DSN-MR          & ResNet-12 & 67.39 $\pm$ 0.82                          & 82.85 $\pm$ 0.56                           \\
DeepEMD          & ResNet-12 & 71.16 $\pm$ 0.87                          & 86.03 $\pm$ 0.58                           \\
FEAT                    & ResNet-12 & 70.80 $\pm$ 0.23                          & 84.79 $\pm$ 0.16                           \\
E$^{3}\text{BM}$     & ResNet-12 & 70.0 $\pm$ n/a                                      & 85.0 $\pm$ n/a                                       \\
APN                 & ResNet-12 & 69.87 $\pm$0.32                           & 86.35 $\pm$0.41                            \\
FRN                     & ResNet-12 & 71.16 $\pm$0.22                           & 86.01 $\pm$0.15                            \\ \midrule
HGNN                                                              & ResNet-12 & \textbf{72.05 $\pm$0.23} & \textbf{86.49 $\pm$ 0.15} \\ \bottomrule
\end{tabular}
\end{table}

\begin{table}[t]
\small\addtolength{\tabcolsep}{-2pt}
\caption{Results on CUB-200-2011}
\label{tab:cub}
\begin{tabular}{lccc}
\toprule
        Methods& Backbone             & 1-shot                          & 5-shot                          \\\midrule
ProtoNet$^{*}$   &Conv4  & 51.25 $\pm$  0.21      &        72.26 $\pm$ 0.18                                                    \\
Adversarial &Conv4& 63.30 $\pm$ 0.94 & 81.35 $\pm$ 0.67 \\ \midrule
HGNN               &Conv4  & \textbf{69.02 $\pm$ 0.22} & \textbf{83.20 $\pm$ 0.15} \\  \midrule

ProtoNet$^{*}$  &ResNet-12   & 68.11 $\pm$  0.21      &        87.33 $\pm$ 0.13                                                    \\
DeepEMD      &ResNet-12     & 77.14 $\pm$ 0.29 & 88.98 $\pm$ 0.49 \\
Neg-Cosine   &ResNet-18     & 72.66 $\pm$ 0.85 & 89.40 $\pm$ 0.43 \\
Centroid &ResNet-18 & 74.22 $\pm$ 1.09 & 88.65 $\pm$ 0.55 \\ \midrule
HGNN               &ResNet-12   & \textbf{78.58 $\pm$ 0.20} & \textbf{90.02 $\pm$ 0.12} \\  \bottomrule

\end{tabular}
\vspace{-0.4cm}
\end{table}


\subsection{Datasets and Settings}
\paragraph{Datasets} Three widely used FSL benchmarks, \textit{Mini}ImageNet \cite{vinyals2016matching}, \textit{Tiered}ImageNet \cite{ren2018meta} and CUB-200-2011 \cite{wah2011caltech} are used in our experiments. \textit{Mini}ImageNet contains a total of 100 classes and  600 images per class. We follow the standard splits provided in ~\cite{vinyals2016matching}, consisting of 64 classes for training, and 16 classes and 20 classes for validation and testing respectively. \textit{Tiered}ImageNet is a larger subset of the ImageNet ILSVRC-12, comprising 779,165 images from 608 classes. They are divided into 351, 97, and 160 classes for training, validation and testing respectively~\cite{chen2018closer}. Different from the other two datasets, CUB-200-2011 is a fine-grained classification dataset. It includes 11,778 images from 200 different bird classes. The 200 classes are divided into 100, 50, 50 classes for training, validation and testing respectively as in \cite{liu2020negative,afrasiyabi2019associative}.  In all datasets, images are downsampled to $84 \times 84$ as per standard.

\begin{figure*}[ht]
    \centering
    \subfigure[PGNN]{\includegraphics[width=0.48\textwidth]{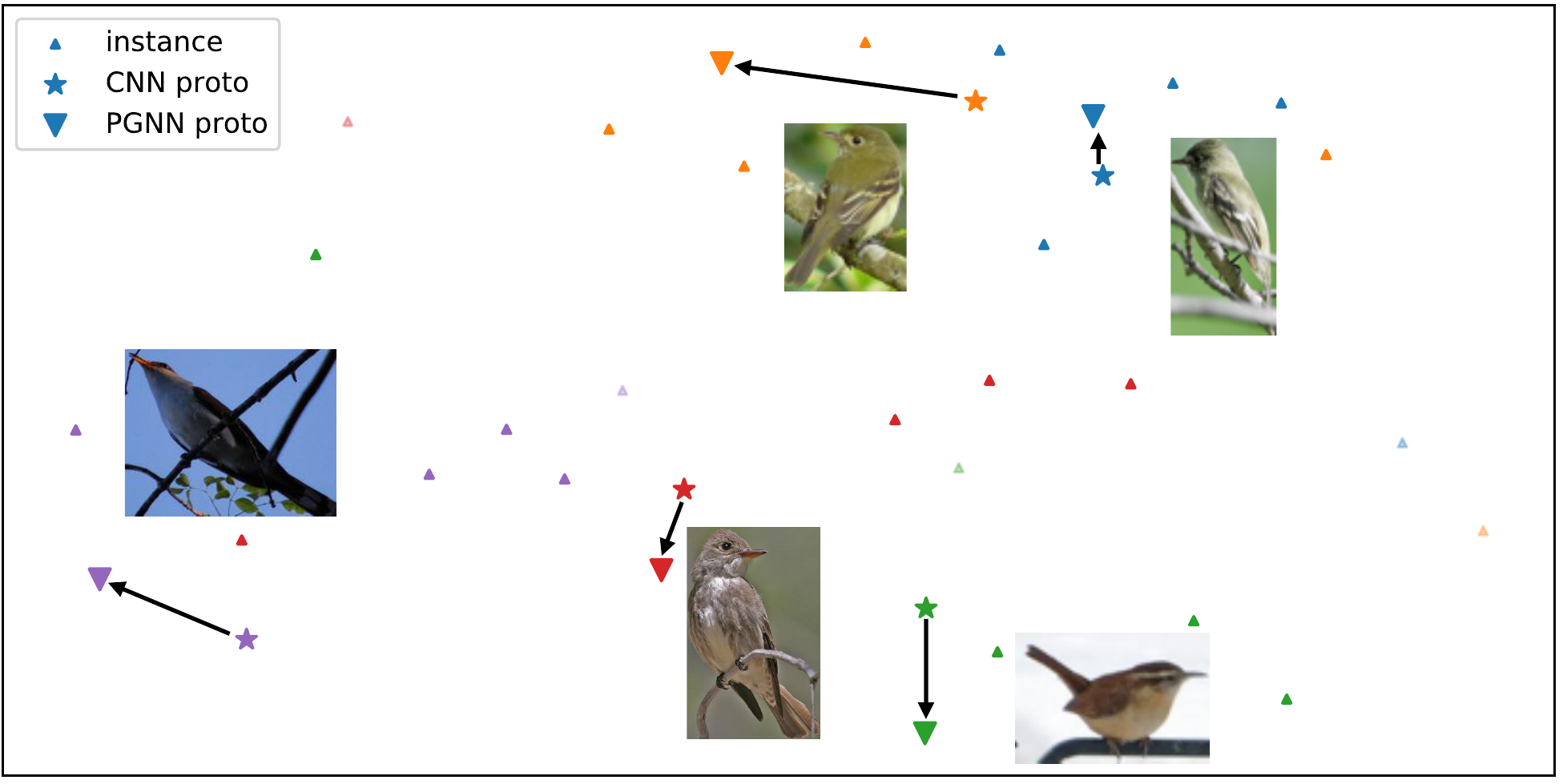}} \label{fig:vi_pgnn}
    \subfigure[IGNN]{\includegraphics[width=0.48\textwidth]{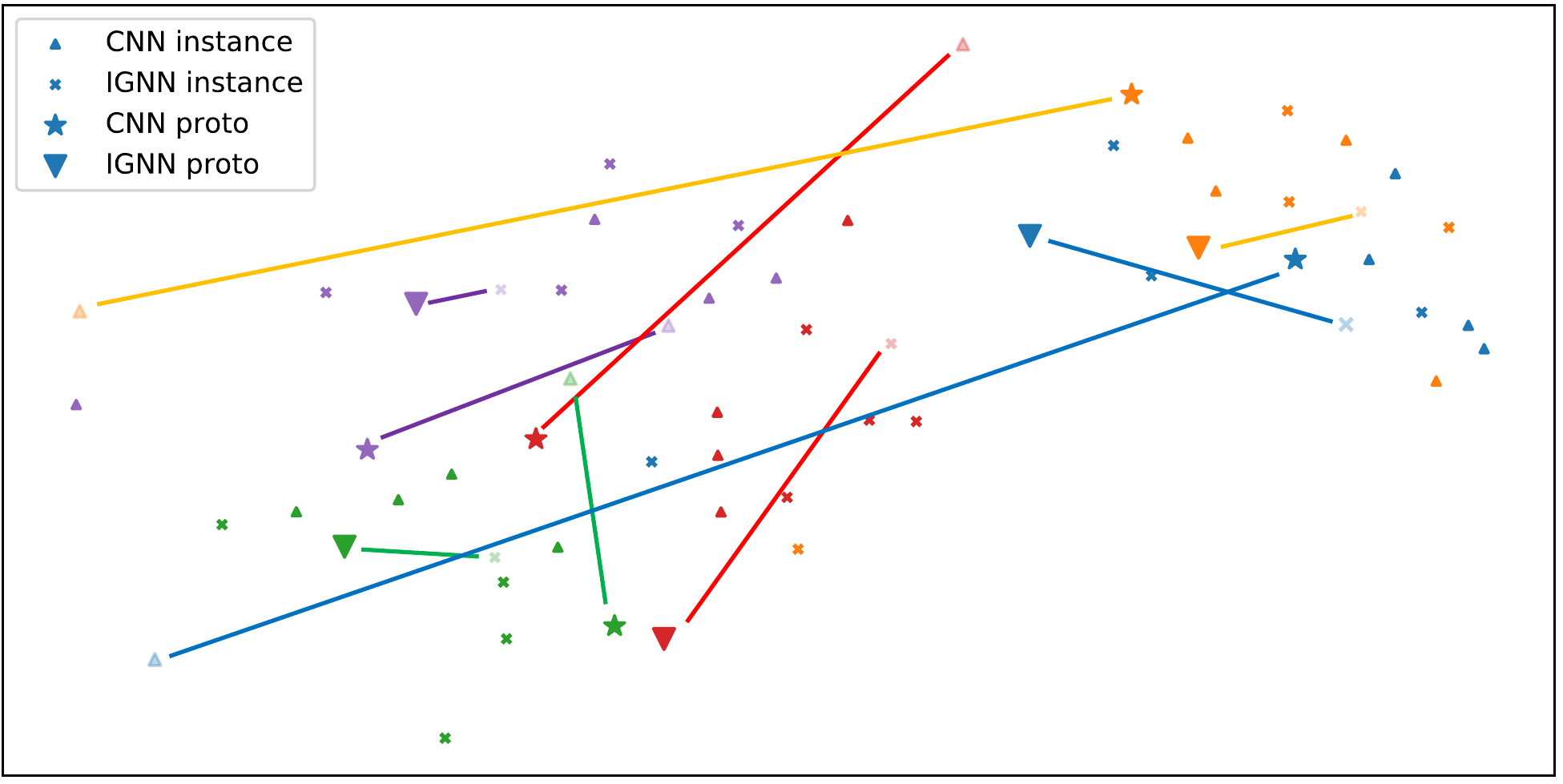}} \label{fig:vi_ignn}
    \caption{Qualitative results on PGNN and IGNN alleviating the class overlapping and outlier issues respectively.  The t-SNE projection of instances and prototypes of 5 classes in CUB-200-2011 are shown here.  In (a), we use vectors to indicate how class prototypes moves from before (CNN protos) to after PGNN updates (PGNN Protos). It can be seen that after the PGNN update, the 5 class prototypes are clearly more separable.  In (b), the distance between an outlying instance and its class prototypes (before and after IGNN updates) are highlighted using color-coded straight lines (same color means the same outlier). IGNN clearly neutralizes the negative  impact of outliers by pulling those outliers closer to their prototypes, indicated by the shorter lines to IGNN protos.  }
    \label{fig:visualisation}
\end{figure*}

\paragraph{Feature Embedding Network}
As in many other CNN-based visual recognition tasks, a feature embedding network is required in a FSL model and the choice of its backbone has a major impact on its performance. For fair comparisons with prior works, two widely used backbones are adopted in our experiments, namely Conv4 and ResNet-12. Following \cite{snell2017prototypical}, the Conv4 backbone has four convolutional blocks and its final output feature dimension is 64. The ResNet-12 backbone is used in most of the state-of-the-art models  \cite{zhang2020deepemd,ye2020few,liu2020negative}. It consists of four residual blocks, and the output feature dimension is 640. Following the common practice \cite{liu2020negative,ye2020few,zhang2020deepemd}, we pre-train our feature embedding network with supervised learning on the whole training set before the episodic meta-learning stage.



\paragraph{Baselines}
Three types of baselines are chosen for comparisons: (1) representative FSL methods including MAML~\cite{finn2017model} (optimization-based), ProtoNet~\cite{snell2017prototypical} (embedding-based), FEAT~\cite{ye2020few} (task-adaptation), and Distill~\cite{tian2020rethinking} (without episodic meta-learning), (2) GNN-based methods including GNN~\cite{garcia2017few}, EGNN~\cite{kim2019edge}, TPN~\cite{liu2019TPN}, DPGN~\cite{yang2020dpgn} and BGNN~\cite{luo2019learning}, and (3) the state-of-the-art (SOTA) methods published in 2020 and 2021, including Centroid~\cite{afrasiyabi2019associative}, Neg-Cosine~\cite{liu2020negative}, DSN-MR~\cite{simon2020adaptive},
DeepEMD~\cite{zhang2020deepemd}, E$^{3}\text{BM}$~\cite{liu2020ensemble}, 
MABAS~\cite{kimmodel}, APN~\cite{luattentive}, Adversarial~\cite{afrasiyabi2019associative}, ArL~\cite{zhang2021rethinking}, PSST~\cite{chen2021pareto}, FRN~\cite{wertheimer2021few}.

\subsection{Main Results}
The comparative results on \textit{Mini}ImageNet, \textit{Tiered}ImageNet  and CUB-200-2011 are shown in Tables \ref{tab:mini}, \ref{tab:tiered} and \ref{tab:cub} respectively. The following observations can be made. \textbf{(1)} Our HGNN achieves the new SOTA on all three datasets under both the 1-shot and 5-shot settings and with both the Conv4 and ResNet-12 backbones, validating its effectiveness. \textbf{(2)} Our GNN-FSL model significantly outperforms all five existing GNN-FSL models \cite{garcia2017few,luo2019learning,liu2019TPN,kim2019edge,yang2020dpgn} under the inductive setting\footnote{For fair comparison, we use the same backbone pre-trained on the base training dataset. 
}. This verifies our hypothesis that jointly meta-learning a GNN-based label propagation/classification module with a feature embedding network confuses the objectives of rapid learning and feature reuse. This results in inferior performance under the inductive setting where, without the access to the full query set, the usefulness of  label propagation is limited. \textbf{(3)} Overall, the advantages of our HGNN over the SOTA alternatives under the more challenging 1-shot setting and with the fine-grained CUB-200-2011 dataset are more pronounced. This is expected: our IGNN and PGNN are designed to address the badly sampled shots problems in the support set. With fewer shots and more inter-class overlapping in the fine-grained cases, these problems are more acute and hence the clearer advantages of our HGNN.


\subsection{Are PGNN and IGNN Doing Their jobs?} Our PGNN and IGNN are designed to solve  the inter-class overlapping and  outlier issues caused by badly sampled instances in each few-shot classification task. 
These two issues severely impact the performance of a few-shot learning method, but are rarely discussed before.  Figure~\ref{fig:visualisation} visualizes how the feature embedding distributions of the  support set instances in a task sampled from CUB-200-2011 under the 5-way 5-shot setting. This qualitative result aims to show how the 5 prototypes and outlying instances are distributed before and after the output of the feature embedding CNN is updated by the two GNNs.  We can see clearly that these two GNNs are indeed addressing the two issues as we anticipated: the PGNN pushes the class prototypes further away from each other to tackle the class overlapping problem, while the IGNN produces a more compact intra-class distribution by shortening the distance between the outliers and prototypes, mitigating the outlier problem.

\section{Conclusions}
We have proposed a novel GNN-based FSL model. Different from the existing GNN-FSL methods which utilize GNN as a label propagation tool to be jointly meta-learned with a feature embedding network, we argue that a GNN is best used in FSL as a feature embedding task adaptation module. In particular, it should address the outlying samples and class overlapping problems commonly existing in FSL through the task adaptation. To that end, an instance GNN and a prototype GNN are formulated and their complementarity is exploited in a hybrid GNN framework. Extensive experiments demonstrate that our HGNN is indeed effective in addressing the poor shot sampling problems, yielding new state-of-the-art on three benchmarks. 

\bibliography{aaai22}

\clearpage

\appendixpage
\setcounter{table}{0}  

\renewcommand{\thetable}{A\arabic{table}}

\section{Experimental Settings}

\subsection{Implementation Details}

Our model is implemented using PyTorch.  We use the Adam \cite{kingma2014adam} optimizer to train our model when Conv4 backbone is used with an initial learning rate of $1e^{-4}$ and $1e^{-3}$ for the feature embedding network  and HGNN respectively. In the experiments using ResNet-12 as the backbone, a SGD optimizer is used, and the initial learning rate is set to $2e^{-4}$ and $2e^{-3}$ for the backbone and HGNN respectively. We train our model with 200 epochs for all experiments, each epoch including 100 episodes. We adopt step-wise learning rate decay and halve the learning rates every 40 epochs in all experiments. The code and trained models will be released soon.

\subsection{Evaluation Protocols}  
We evaluate our model for the 5-way 1-shot/5-shot tasks under the inductive setting as in most existing works. During testing, we sample 15 queries for each class per task. We test our model with 10,000 randomly sampled tasks from the testing datasets, and report the top-1 mean accuracy as well as the 95\% confidence interval. 

\section{Ablation Study}

\subsection{Contributions of Each GNN} HGNN is composed of a PGNN and an IGNN integrated with a ProtoNet \cite{snell2017prototypical}. In this experiment, we investigate the contributions of each GNN. The results are shown in Table~\ref{tab:ablation1}. It can be seen clearly that: (1) Task adaptation on the feature embedding learned using ProtoNet brings improvements with both PGNN and  IGNN -- around 2\% boost in accuracy is achieved. The novel PGNN is the strongest of the two.  (2) When the two GNNs are combined in our HGNN, a further boost is obtained, indicating their complementarity. The boost is of small absolute value. However,  given the fact that the results on MiniImageNet have saturated recently
at around 82\%, this improvement is still significant.

\begin{table}[t]
    \centering
    \caption{Ablation results on 5-shot \textit{Mini}ImageNet}
    \small
    \begin{tabular}{cccc|c}
    \hline
    ProtoNet  & PGNN & IGNN & HGNN &\textbf{Acc} \\
    \hline
    \checkmark& & & & 80.49  \\
    \checkmark & \checkmark & & & 82.62\\
    \checkmark &  & \checkmark & & 82.20  \\
    \checkmark & &  & \checkmark & \textbf{83.00}\\
    \hline
    \end{tabular}
    \vspace{-0.4cm}
    \label{tab:ablation1}
\end{table}

\subsection{Different Consistency Losses}
\begin{table}[t]
\centering
\caption{Effects of different consistency losses on \textit{Mini}ImageNet}
\label{tab:ablation2}
\begin{tabular}{l|cccc}
\toprule
      &None & L1    & MSE   & KL    \\ \midrule
1-shot &66.70& 66.82      &  66.75     & \textbf{67.02}     \\
5-shot &82.61& 82.71 & 82.65 & \textbf{83.00} \\ \bottomrule
\end{tabular}
\vspace{-0.2cm}
\end{table}
In our HGNN, the IGNN and PGNN are trained jointly with a consistency loss based on KL divergence. Alternative consistency loss functions exist, including L1 loss, mean squared error (MSE). Table~\ref{tab:ablation2} shows that adding the consistency loss helps and the KL divergence based loss is the best option in this case.  
\subsection{Different Adjacency Operators}
\begin{table}[t]
\centering
\caption{5-way 5-shot results with different adjacency operators. `C' denotes concatenation and `S' represents subtraction.
}
\label{tab:attention}
\small
\begin{tabular}{l|ccccc}
\hline
     & inner product & C+FC & C+Conv & S+FC & S+Conv \\ \hline
IGNN &     82.21        &      82.34     &     82.41        &      82.30          &       82.39           \\
PGNN &      82.61       &       82.35    &      82.22       &      82.47          &     82.07             \\
HGNN &     \textbf{83.00}        &       82.61    &    82.93         &     82.64           &         \textbf{83.01}         \\ \hline
\end{tabular}
\end{table}
The adjacency operator in Equation 2, which computes the correlation between different nodes, is a key component in a graph's operation. Many different operators have been considered before. These include a concatenation or subtraction between two nodes' features followed by a linear layer or convolutional layers, or a simple non-parametric inner product as adopted in our GNNs.  Table~\ref{tab:attention} shows that  different adjacency operators have little impact on the performance. We thus choose the simplest  inner product adjacency operator in our HGNN owing to its memory efficiency.

\subsection{Effect of graph depths}
Prior works~\cite{kim2019edge,luo2019learning} on graph have shown that a deeper graph would achieve better performance. In this section, we investigate how the depth of IGNN and PGNN would affect the performance. The results in Table~\ref{tab:layer} suggest that deeper GNNs yield worse results. As discussed in \cite{li2018deeper}, one possible reason is that a GNN creates Laplacian smoothing, where the node's feature will become more and more similar as the depth increases. Such an effect can have a detrimental impact on the FSL performance.

\begin{table}[t]
\centering
\caption{5-way 5-shot results on \textit{Mini}ImageNet with different GNN depths.}
\label{tab:layer}
\small
\begin{tabular}{l|ccc}
\hline
        & 1 layer & 2 layers & 3 layers \\ \hline
IGNN & \textbf{82.20}  &  81.57   &  78.08  \\
PGNN &  \textbf{82.62}  &  81.71   &  81.38  \\\hline
\end{tabular}
\vspace{-0.4cm}
\end{table}

\subsection{Results under the transductive setting}

Different with the inductive setting, where each query sample is predicted independently from other queries, all the queries are simultaneously processed and predicted together under the transductive setting. Generally, for graph neural network methods, $N$ queries are formed in one group and processed together in the $N$-way setting. However, in the previous works, the order of the queries are fixed when processing these queries during training and testing. For example, under the transductive 5-way setting, the query group are always fed into the model with the label order $[0,1,2,3,4]$. The fixed order is impossible in real scenarios as we don't know the label of testing queries. So we suppose that the right setting should shuffle the queries in one group and may have multiple instances belonging to one class. 

Table~\ref{tab:order} shows the result of GNN-based methods (GNN~\cite{garcia2017few}, EGNN~\cite{kim2019edge}, DPGN~\cite{yang2020dpgn}, BGNN~\cite{luo2019learning}) under the transductive setting with Conv4 backbone on \textit{Mini}Imagenet. We find that shuffling the queries during inference has a detrimental impact and results in much worse results. The main reason is that the trained model remembers the label order of queries in each group rather than learning the instance-class relationship.  For our model, two branches PGNN and IGNN can still propagate message among the nodes. Specifically, PGNN is used by the same way with the inductive setting while IGNN still propagates the information from the labeled image feature to the query image feature. The only different implementation of our model under the transductive setting is that we allow the query images to propagate information between each other so that they could make full use of the aggregated information to increase the inter-class distance and decrease the intra-class distance. From Table~\ref{tab:order}, we can see that our HGNN surpasses all other GNN-based methods for FSL under the corrected transductive setting, which validates the effectiveness of our model.

\begin{table}[t]
\centering
\caption{Results on the \textit{Mini}ImageNet dataset using a Conv4 backbone under the transductive settings. }
\label{tab:order}
\begin{tabular}{c|cc|cc}
\toprule
     & \multicolumn{2}{c|}{Fixed order} & \multicolumn{2}{c}{Shuffled order} \\ \midrule
          & 1shot        & 5shot       & 1shot          & 5shot         \\ \midrule
GNN    &                   48.68     &      69.37        &       20.10          &         19.98       \\
EGNN  &      59.74         & 76.87        & 46.53           & 57.09          \\
DPGN &     66.01         & 82.83        &       20.57          &      21.22          \\
BGNN  &          91.22         & 93.52        & 54.70           & 69.11          \\ \midrule
HGNN &          -        & -        &\textbf{55.60}           & \textbf{72.52}          \\ \bottomrule
\end{tabular}
\end{table}

\end{document}